# Flipping the Switch on Local Exploration: Genetic Algorithms with Reversals

Ankit Grover[1,2], Vaishali Yadav[2] , Bradly Alicea[1,3]

1 Orthogonal Research and Education Laboratory, Champaign-Urbana, IL and Worldwide
2 Manipal University Jaipur, Rajasthan, India
3 OpenWorm Foundation, Boston, MA

```
Email: agrover112@gmail.com
```

**Abstract:** One important feature of complex systems are problem domains that have many local minima and substructure. Biological systems manage these local minima by switching between different subsystems depending on their environmental or developmental context. Genetic Algorithms (GA) can mimic this switching property as well as provide a means to overcome problem domain complexity. However, standard GA requires additional operators that will allow for large-scale exploration in a stochastic manner. Gradient-free heuristic search techniques are suitable for providing an optimal solution in the discrete domain to such single objective optimization tasks, particularly compared to gradient-based methods which are noticeably slower. To do this, the authors turn to an optimization problem from the flight scheduling domain. The authors compare the performance of such common gradient-free heuristic search algorithms and propose variants of GAs. The Iterated Chaining (IC) method is also introduced, building upon traditional chaining techniques by triggering multiple local searches instead of the singular action of a mutation operator. The authors will show that the use of multiple local searches can improve performance on local stochastic searches, providing ample opportunity for application to a host of other problem domains. It is observed that the proposed GA variants have the least average cost across all benchmarks including the problem proposed and IC algorithm performs better than its constituents.

## 1 Introduction

In a wide range of complex systems, there are several problem domains where substructure, noise, and local variability are prevalent. These properties give rise to a very rough landscape, with many local minima. Genetic algorithms (GAs) can provide a means to navigate such landscapes, but standard operators such as mutation and recombination limit the ability to escape local minima without destroying accumulated search information.

Fortunately, a mechanism has been identified that is often used at the phenotypic or epigenetic levels of evolutionary systems: switching. Switching is the transition from one state to another by means of a quick, abrupt change. Switching can be characterized in forms such as electrical switches, molecular triggers, step functions, and first-order phase transitions. One way in which switching is useful for computational environments is in modeling switching as a computational universal phenomenon. For example, cellular automata [1] to understand how the implementation of rules can produce complex patterns that approximate those seen in biological systems. When applied in parallel, these rules result in switching points at the macro-level, which introduces heterogeneity in the pattern. More relevant to the kinds of switching observed in evolutionary systems, adaptive behaviors enabled by switching and found in the lac operon system, logic circuits, and RNA phenotypes. In two models of such systems, innovation via mutation (lac operon, [2]) and increased speed of evolutionary innovations (logic gates and RNA phenotypes, [3]) serve as examples of how switching can be encoded in a computational framework.

Historically, chaining methods are used in a wide variety of algorithms and can be applied to many different types of problem domains. One example of this is in applications to multiple genome comparisons. In this problem domain, this is done by constructing a maximum weighted path on a weighted directed acyclic graph [4]. Optimal solutions to this type of problem [5, 6] allows for spatial and temporal specificity, which decomposes the problem domain to a series of local minima. Chaining is also used for optimization in a number of problem domains, including nearest-neighbor, lexical, and backwards [7]. Lexical chaining is also related to word-sense disambiguation [8], which uses a switching mechanism in a linguistic context. This problem poses switching as a means of discrete phase change and lends itself to exploring subdomains of a problem.

## 2      Literature Review and Methodology

### 2.1      Literature Review

A flight scheduling problem can be used to demonstrate the efficacy of our approach. The fliscopt approach utilizes a dataset of schedules consisting of features such as flight cost in dollars, departure, arrival times, city of origin, destination. This makes our problem NP-Hard [9] and heuristic algorithms with optimal solutions are preferred. Our dataset consists of six-city names and airport abbreviations [10]. A single person is assumed to be present at each city of origin. The domain D in this case is a discrete vector v(0,9) of magnitude 12, representing round trips for the six-city example. The final schedule must take this condition into account. The search space is of the order of $10^{12}$, since 10 possibilities exist for each of the 12 trips. From this, a cost function can be created which can efficiently create flight schedules from origin cities to a particular destination city with least total price and total wait times.

Flight scheduling is often closely related to fleet assignment [11]. The authors [12] use a Lagrangian relaxation along with sub-gradient methods. Similarly, the authors in [13] use a MAGS which is based on the Ant Colony Optimization Method [14]. Simple heuristic search techniques are used in lieu of gradient-based information, while alternate variants of a standard GA are implemented. Some of these variants have been used by [15] for image encryption, also for Generation of S-Boxes [16] and by authors [17] using Reverse Hill Climbing for the Busy Beaver Problem [18], the authors in [19] use a Binary Differential Evolution algorithm for airline revenue management which is NP-Hard problem as well. Similarly, the authors of [20] use a GA with 2-Dimensional mutation and crossover operations for Aircraft Scheduling. The Nevergrad software package [21] uses a similar non-gradient based principle with single iteration of chaining, however our method involves the use of mutation operators and multiple iterations. By comparison, early stopping criteria combined with mutations are used to influence local minima. Also, our second proposed variant is similar to [22] which uses Tournament Selection instead and has been applied to solve the parking lot path optimization problem [23]. GAs with reversals have also been by for the Scary Parking Lot problem.

### 2.2  Defining the Fitness Function

In the following problem, there is a need to meticulously define the cost (fitness) function [24] such that it takes into account both the cost of flights (dollars) but also the waiting times between different flights. The wait times need to be minimized and so does the cost. Thus, the final function is an additive of both quantities. The cost (fitness) function penalizes cases where flights are too far apart.  The function get_minutes is used to convert hours into minutes. The time complexity of our fitness function is O(L) or O(N), where L is the length of the solution.

Benchmark functions are chosen such that it would be difficult to converge easily. This was done by choosing functions such that they followed certain properties [25, 26] of high dimensionality, since most real-world problems are multi-dimensional in nature; they were non-separable, continuous, convex, and unimodal. Average cost($\overline{cost}$), Standard deviation($\sigma$), number of function evaluations(n.f.e), and runtime (ms) are used as metrics for evaluation.

### 2.3  Switching Mechanism

The idea of Switching, which is essentially an sudden  switch(change) in behavior [27],  discusses how such changes can be  are incorporated in living organisms for improving their fitness. Such switching behaviors can be utilized in algorithms as well, for improving fitness. Two types of switching are utilized here: intra-algorithmic and inter-algorithmic. For intra- algorithmic switching, the change in the phenomenon takes place inside the algorithm itself. In the case of inter- algorithmic switching, the switching mechanism occurs between algorithms, rather than within a single algorithm. The authors compared the following algorithms: Simulated Annealing (SA), Random Search (RS), Hill Climbing (HC), standard GAs, and their proposed variants: GA with Reversal Genetic Operations, GA with Reversals, GA with Stochastic Search Reversal, and the IC method.

In this algorithm, a GA in which the order is reversed is used to perform the genetic operations of mutation and crossover (see Figure 1). Thus, the GA consists of an Elitist selection step followed by crossover and then a mutation step. The probability of mutation $P_{mutation}$ thus, becomes the probability of crossover. The GA with Reverse Operations converges a bit faster to optimal cost.

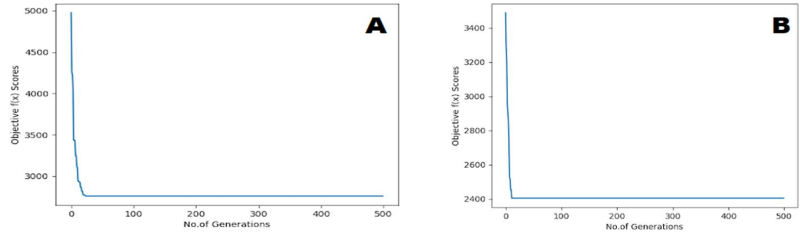

**Fig. 1.** Plot for a standard GA (A), and for a GA with Reverse Operations (B) using the same seed for a single run.

This method can be understood as a coarse approximation of the Differential Evolution method without the differential weight F and selection step at the start of the iteration. Being an approximation, this method provides some increase in performance when GA fails. Therefore, Figure 1 demonstrates results on Problem function that simply reversing operations isn't enough. To potentially overcome this limitation, a reversal strategy is described in the next section.

**GAs with Reversals.**

GA reversals are a form of switching which aims to introduce genetic diversity by means of reversing the maximization or minimization step. Since this reversal happens within the same GA it is a form of intra-algorithmic switching. Our idea is somewhat similar to [28] two neural networks where one tries to minimize whereas the other tries to maximize the other in a zero-sum game. Here the competition happens within the optimization algorithm itself.

Elitist selection is useful when the algorithm can get stuck in the loss landscape of functions with many local minima and is slightly difficult to converge. This has been done by using the reversal step. Minimizing for the objective, the objective function is maximized for a fixed number of steps defined by the $step_{length}$ parameter, and the $num_{reversals}$ is controlled by the $n_k$ parameter using Equation 1. Reversals are performed for all iterations other than the first iteration. The iteration at which the reversal starts is the iteration i divisible by $n_k$. By using such a short reversal process, local minima can easily be escaped. This forces the algorithm to find better search spaces.

$$num_{reversals} = \frac{num_{generations}}{n_k} \quad (1)$$

$Where\ num_{generations} \in Z;\ n_k \in N\ \&\ n_k \leq num_{generations}\ \forall\ i \in num_{generations}\ \&\ i \neq 0.$

**GAs with Stochastic Reversals**

While an exemplar random search algorithm is implementable, performance on benchmarks may yield poor performance. Thus, GAs with stochastic reversals can be used (Figure 2A), in which the algorithm performs a local random search in reverse objective maximization mode similar to how it does for GA with Reversals. This is shown in comparison to GAs with Reversals (Figure 2B), which itself demonstrates performance improvements compared to a standard (vanilla) GA. Figure 2A clearly shows how a reverse Stochastic Search perturbs the solution space, leading to worse solutions during the reversal process. In this mode the GA provides diversity in solutions by using Random Search instead. A consequence of this might be it reaching arbitrary search spaces in the landscape while also escaping local minima.

### 2.4 IC Algorithm

Solution initialization can be used in heuristic algorithms, which provides any algorithm a good starting point in the solution space. This involves providing an initial starting population/solution to a GA such as Random Search (RS) and Hill Climbing (HC). IC can be simplified to an Iterated Local Search by using two similar algorithms and removing the operators between them [29, 30]. Prior knowledge is used in the form of solution initialization. Full code for the IC algorithm can be found in [24]

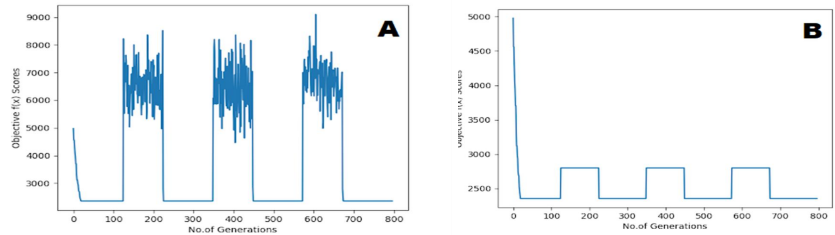

**Fig. 2.** Random Search Reversal (A), and GA with Reversals (B).

For the purposes of this paper, the initial solution will be referred to as weights, initialized weights, or solution weights interchangeably. This is the convention used in deep learning and suits this context as well. In the following scenario, solution initialization is a discrete approximation of weight initialization in neural networks. Major difference being the initialized weights are from a pre-trained algorithm.

Prior knowledge plays an important role in deep learning. This is especially true when the model is based on both prior knowledge and temporal recurrence. In such cases, a deep learning algorithm can outperform those without pre-training [31, 32]. A GA-based method will be used where instead of a single initialization, re-initialization is iteratively performed. GAs use mutation and crossover, which prevent stagnation and locally improve the solutions.

In our meta-algorithm, two algorithms are used to transfer its best solutions to one another over a certain number of rounds. Each algorithm learns from the other to a certain extent. The parameter *rounds* is analogous to epochs in training machine learning algorithms. Each round consists of running the Initial algorithm and the Chained Algorithm which uses weights from above. The Chained weights are then recursively implemented in the Initial algorithm. For the final iteration of the meta-algorithm (*rounds* -1), weights are passed directly to the Chained instead until terminating.

## 3  Further Methodological Considerations

### 3.1  Overcoming Drawbacks with IC

One of the major issues of using such an initialization scheme is to account for Initial algorithm's solution starting to deteriorate, and thus providing bad weights to the Chained Algorithm as a result decrease the global cost. Running the algorithm for a large number of rounds only aggravates the issues. Parameters are required which can quickly prevent divergence of the global cost during a deteriorating solution. Global cost in the following context refers to the cost of the meta-algorithm itself and cost refers to the local cost of the constituent algorithms. Our goal is to thus minimize the global cost. Thus, an Early Stopping mechanism such as that used in Deep Neural Networks controls the global cost [33].

The authors introduce two parameters $n_{obs}$: the number of observations over cost is averaged, and *tolerance* : the amount up to which a divergence of global cost is tolerated (see Equation 2). A larger value in $n_{obs}$ should handle divergence much more. In the given equation the right-side calculates the average of the global cost up to $n_{obs}$ and the left-side the *tolerance* tries to minimize the current cost by subtracting a factor. If the cost is substantially higher than the $n_{obs}$ (despite the *tolerance* factor), the meta-algorithm terminates early, preventing waste of resources. A default value of *tolerance* of 90 and $n_{obs}$ of 2 were used. Very high values of *tolerance* cause termination in early rounds

A demonstration of the Early Stopping mechanism and the consequences of limited diversity are shown in Figure 3.

$$cost - R > \sum \frac{scores[\,-n_{obs}:]}{n_{obs}} \quad \rightarrow \quad best_{soln}, last_{score}, scores\ and\ n.f.e \quad (2)$$

Where $R \in Z$, $tolerance \leq R \leq 100$, from a discrete U distribution, $scores[-n_{obs}:]$ represents last n cost observations. This method has some drawbacks. One notable consequence is a lack of transfer learning from one cost landscape to another. The model depends on the quality of the weights, here the performance is determined by how optimized our initialized weights are with respect to our minimization problem. Therefore, it is important to have good weights. Another drawback is that the initial algorithm being weak, can often perturb the results to a degree that the Chained algorithm then fails to further reduce the cost and thus our improvements converge without providing a benefit with respect to an optimal solution. GAs were observed to converge the most with respect to Chained algorithms. A similar phenomenon is observed with two similar algorithms: each algorithm converges simultaneously, and this leads to the meta-algorithm being stuck as shown in Figure 3. To overcome the lack of diversity in these representations, the OnePointMutation operator [34] is used. A mutation is introduced randomly after each Initial algorithm, while a definite mutation to the solution of the Chained algorithm. This provides the necessary diversity required and prevents the algorithms from being stuck and explores the search space better. Introducing mutation greatly decreases the $\overline{cost}$, of our problem. The final performance of this method depends purely on the algorithms chosen and the problem at hand, therefore there might be cases where our Itertated Chaining method might fail.

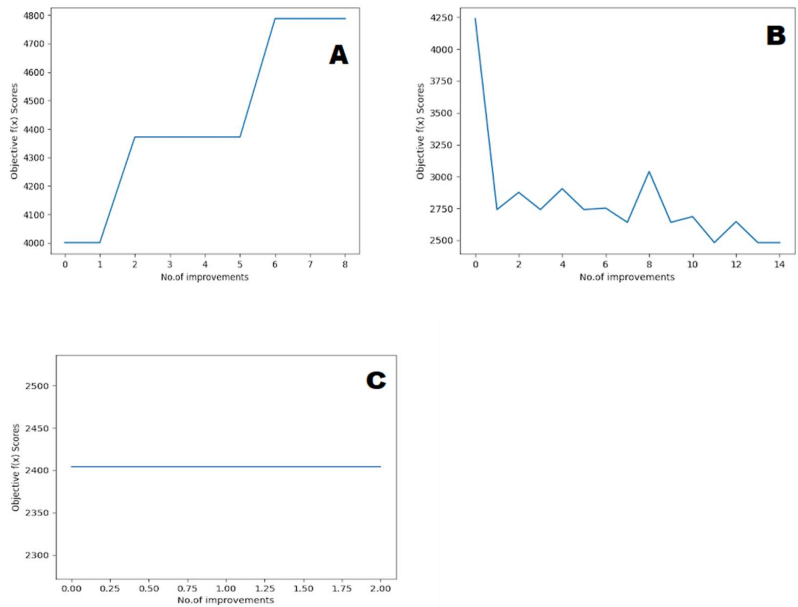

**Fig. 3.** (A-C) A: A demonstration of divergence in an IC algorithm (no stopping mechanism - converges towards a higher score). B: After applying Early Stopping criteria (converges

towards a lower score). C: A GA-GA Iterated Chain that fails to provide any diversity and converges without mutations.

### 3.2 Hardware and Performance Issues

All the experiments were carried out on a pc with Windows 11 using Windows Subsystem for Linux , using an Intel(R) Core(TM) i7-9750H CPU with 2.60GHz, 2592 MHz, 6 Core(s) and 12 Logical Processor(s) with a RAM of 32.0 GB with available physical memory of 16.5 GB. It is to be noted that since experiments were performed on Windows Subsystem for Linux(WSL) 2 , the wall clock time on a native Linux distribution might be faster.

Background processes running on the system will also have a direct impact on the wall clock time. Wall clock time will be referred to as run-time. To further decrease the run-time the authors chose the PyPyv7.3.5 [35] interpreter instead of the native CPython implementation of Python along with asynchronous multiprocessing on all the cores.

## 4 Results

### 4.1 Experimental Setting

In the following section we describe the parameters used in our experiments. For proper evaluation we set an initial seed for generating the population. This seed was chosen randomly, and results reported accordingly as the average of 20 runs with different seeds. For Hill Climbing, Simulated Annealing, Random Search the maximum iterations are capped at 100, as no further improvements are noticed beyond 100. Simulated Annealing had an initial temperature set to 50000, a cooling rate of 0.95 and step size in either direction as 1. The Genetic Algorithm variants have a set of common parameter values which are population size set to 100, generations set to 500 and Elitist selection rate of 0.2. The Genetic Algorithm with Reversal introduces 2 new parameters which are $num_{reversals}$ set to 1 and $step_{length}$ set to 100. The number of reversals can be a minimum of 1 and always a positive value. The Step Length again is always a positive value . The Iterated Chaining algorithm also has a set of parameters used to control the local searches involved in them, the $rounds$ is set to 1, the $tolerance$ is set to 90 and $n_{obs}$ is set to 2.

### 4.2 Results and Observations

The run-time and descriptive statistics for single runs of our various candidate algorithms are reported in Table 2. In general, increasing the number of iterations (as was conducted for Hill Climbing, RandomSearch, and Simulated Annealing algorithms) demonstrates that increased run-time results n.f.e (number of function evaluations) while showing marginal to no improvement relative to all other proposed algorithms. Abbreviations used in Table 2-4 are as follows: **A = standard GA; B =**

**GA with Reverse Operations; C = GA with Reversals; D = Hill Climbing; E = Random Search; F = Simulated Annealing; G = GA with Random Search as Reversals.**

**Table 2**. Run times and descriptive statistics for experimental conditions (different search strategies A-G).

|  | Problem Cost Function | | | | | | | Rosenbrock (13 dimensions) | | | | | | |
|---|---|---|---|---|---|---|---|---|---|---|---|---|---|---|
|  | A | B | C | D | E | F | G | A | B | C | D | E | F | G |
| $\overline{cost}$ | 2780.9 | 2629.8 | 2593 | 4177.7 | 4545.3 | 3726.5 | 2592.9 | 134.55 | 134.35 | 72.5 | 645001 | 209620 | 2.13E+06 | 122.9 |
| $\sigma$ | 205.75 | 213.79 | 183.89 | 817.72 | 271.95 | 578.16 | 168.45 | 194.47 | 186 | 145.85 | 804257 | 90605 | 1.08E+06 | 197.8 |
| $cost_{min}$ | 2356 | 2356 | 2356 | 2759 | 4143 | 2759 | 2356 | 0 | 0 | 0 | 0 | 93099 | 1.15+E5 | 0 |
| $cost_{max}$ | 3081 | 3004 | 2973 | 5839 | 5165 | 4679 | 2888 | 609 | 518 | 510 | 3.03E+06 | 354460 | 3.90E+06 | 609 |
| *n.f.e.* | 1000 | 1000 | 1099 | 328 | 100 | 512 | 1099 | 1000 | 1000 | 1099 | 1093 | 100 | 512 | 1099 |
| *runtime* in *ms* | 9.36 | 9.66 | 10.16 | 0.33 | 0.17 | 0.24 | 9.97 | 0.1 | 0.1 | 0.3 | 0 | 0 | 0 | 0.2 |

**Table 3.**

|  | Zakharov (13 dimensions) | | | | | | | Griewank (13 dimensions) | | | | | | |
|---|---|---|---|---|---|---|---|---|---|---|---|---|---|---|
|  | A | B | C | D | E | F | G | A | B | C | D | E | F | G |
| $\overline{cost}$ | 84.5 | 86.3 | 55.3 | 298.7 | 394.1 | 4.03E+08 | 58.6 | 15.7 | 32.1 | 29.9 | 24 | 178 | 385.3 | 33.8 |
| $\sigma$ | 26.8 | 36.7 | 23.5 | 110.2 | 385.1 | 5.41E+08 | 32.3 | 7.6 | 18.5 | 15 | 12.4 | 33 | 76.4 | 14.9 |
| $cost_{min}$ | 40.3 | 33.1 | 27.3 | 65.3 | 138.3 | 7.58E+05 | 9.3 | 2.2 | 6.9 | 4 | 8.3 | 110.8 | 275.1 | 11.2 |
| $cost_{max}$ | 159 | 150 | 108 | 505 | 1644 | 2.16E+09 | 136.3 | 36.3 | 77.2 | 62.5 | 47.5 | 228.7 | 600.3 | 67.9 |
| *n.f.e.* | 1000 | 1000 | 1099 | 698.5 | 100 | 512 | 1099 | 1000 | 1000 | 1099 | 78026 | 100 | 512 | 1099 |
| *runtime* in *ms* | 0.1 | 0.1 | 0.2 | 0 | 0 | 0 | 0.2 | 0.1 | 0.1 | 0.2 | 0.2 | 0 | 0 | 0.2 |

By using Chaining (random search and hill climbing), minimum cost is achieved, albeit with a high standard deviation ($\sigma$). However, this is still less than the average of both constituent algorithms. The relative cost is also only 270 greater than that of our GA and significantly less than the $\overline{cost}$, of both. However, the IC algorithm has an obviously high run-time and n.f.e due to the nature of its fundamental operations. The GA and its variants (see Table 4-6) converged completely using Brown(13 dim),

10Booth[36], Ackley_N2 [37], Sphere[38], Three_hump_camel [39], and Schwefel[40] . While these functions were completely minimized, other algorithms (Rosenbrock [41], Griewank [42], Zakharov [43] of 13 dimensions, Schaffer_N1 [43], Matyas [44, 45, 46]) did not. The full results for these functions are not reported.

**Table 4.**

|  | Matyas 2D ||||||| Schaffer_N1 (2 dimensions) |||||||
| --- | --- | --- | --- | --- | --- | --- | --- | --- | --- | --- | --- | --- | --- | --- |
|  | A | B | C | D | E | F | G | A | B | C | D | E | F | G |
| $\overline{cost}$ | 0 | 0 | 0 | 0.4 | 0.1 | 30 | 0 | 0.2 | 0.1 | 0.1 | 0.6 | 0.2 | 1.3 | 0.1 |
| $\sigma$ | 0 | 0 | 0 | 0.9 | 0.1 | 24.4 | 0 | 0.1 | 0.1 | 0.1 | 0.1 | 0.1 | 0.1 | 0.1 |
| $cost_{min}$ | 0 | 0 | 0 | 0 | 0 | 3.1 | 0 | 0 | 0 | 0 | 0.4 | 0 | 0.9 | 0 |
| $cost_{max}$ | 0 | 0 | 0 | 3.9 | 0.5 | 90.3 | 0 | 0.4 | 0.4 | 0.3 | 0.8 | 0.5 | 1.5 | 0.5 |
| n.f.e. | 1000 | 1000 | 1099 | 44 | 100 | 512 | 1099 | 1000 | 1000 | 1000 | 10 | 100 | 512 | 1099 |
| runtime in *ms* | 0.1 | 0.1 | 0.1 | 0 | 0 | 0 | 0.2 | 0.1 | 0.1 | 0.1 | 0 | 0 | 0 | 0.1 |

In most multi-dimensional functions, it is observed our GA variants outperform GAs with better $\overline{cost}$, $\sigma$ with slightly larger run-times, and in terms of n.f.e. Moreover, another thing to note is the $num_{reversals}$ was set at a minimum of 1, showing how a single reversal can drastically help prevent problems of getting stuck in local minima and improve performance. All of our variants, particularly a standard GA with Reversals, can be used in real world problems where the objective function of our problem is complex , multi-dimensional with a very small trade-off for slightly higher n.f.e and run-time.

## 5  Discussion

In this paper, the IC method is applied to a problem domain, which ultimately improves local performance in GAs. When considering the overall effectiveness of IC, performance can be variable, even on the same problem or benchmark.

The IC method was only tested on a single optimization problem (flight scheduling). Other benchmarks were considered, but given the algorithm's dependence on local searches, it would be futile to implement simple benchmark functions. IC outperforms a combination of such local optimization techniques most of the time. Yet reproducibility is difficult, largely but not exclusively due to the high variability between local searches.

# 6     Future Work

The current work is limited to flight scheduling optimization in the discrete domain. As such, the computational experiments were performed for a limited dataset, and more esoteric evolutionary methods such as Ant Colony Optimization (ACO) , Particle Swarm Optimization (PSO) , or Evolution Strategies (ES)  were not fully evaluated. An in-depth investigation of the IC algorithm performance with more local algorithms and problem domains is necessary. Moreover , future evaluations can include benchmarks which have additive noise to reflect real world conditions and shifted minima to check for various biases.